\documentclass[11pt,a4paper]{article}
\usepackage[dvipsnames, svgnames, x11names]{xcolor}
\usepackage[hyperref]{emnlp2020}

\usepackage[utf8]{inputenc} 
\usepackage[T1]{fontenc}    
\usepackage{microtype}
\usepackage{graphicx}
\usepackage{subfigure}
\usepackage{booktabs} 
\usepackage{comment}
\usepackage{hyperref}
\usepackage{mathtools}
\usepackage{multirow}
\usepackage{bookmark}
\usepackage{color}
\usepackage{float}
\usepackage{graphicx}
\usepackage{CJKutf8}
\usepackage{url}
\usepackage{bm}
\usepackage{diagbox}
\usepackage{parskip}
\usepackage{colortbl}
\usepackage{booktabs}       
\usepackage{amsmath, amsfonts, amsthm, amssymb}

\usepackage{amsmath, amsfonts}

\usepackage{times}
\usepackage{latexsym}

\definecolor{mygray}{RGB}{224,224,224}

\usepackage{newtxtext,newtxmath}

\aclfinalcopy 

\title{Neural Semi-supervised Learning for Text Classification \\
Under Large-Scale  Pretraining 
}
\author{
Zijun Sun$^\clubsuit$, Chun Fan$^{\spadesuit\bigstar}$, Xiaofei Sun$^\clubsuit$, Yuxian Meng$^\clubsuit$, Fei Wu$^\blacklozenge$ and Jiwei Li$^{\blacklozenge\clubsuit}$\\
  $^\blacklozenge$Department of Computer Science and Technology, Zhejiang University\\
  $^\spadesuit$Computer Center of Peking University, 
  $^\bigstar$Peng Cheng Laboratory\\
  $^\clubsuit$ Shannon.AI\\
  \{zijun\_sun, xiaofei\_sun, yuxian\_meng, jiwei\_li\}@shannonai.com\\
  fanchun@pku.edu.cn,
  wufei@zju.edu.cn
}

\date{}

\begin{document}
\maketitle

\begin{abstract}
The goal of semi-supervised learning is to utilize the unlabeled, in-domain dataset $U$ to improve models trained on the labeled dataset $D$.   
 Under the context of 
 large-scale language-model (LM) pretraining, 
 how we  can make the best use of
  $U$   is poorly understood: 
 Is semi-supervised learning still beneficial 
 with the presence of  large-scale pretraining?
 Should $U$ be used for in-domain LM pretraining or pseudo-label generation?
How should the pseudo-label based semi-supervised model 
  be actually implemented?
How different semi-supervised strategies (e.g., self-learning) affect performances regarding $D$ of different sizes, $U$ of different sizes, etc. 

In this paper, we conduct comprehensive studies  on  semi-supervised learning in the 
task of text classification 
 under the context of 
large-scale
LM pretraining.
Our studies shed
important
 lights on the  behavior of semi-supervised learning methods. 
 We find that: 
(1)  with the presence of  in-domain LM pretraining  on $U$, open-domain LM pretraining \cite{devlin2018bert} 
is unnecessary, and we
are able to achieve better performance with pretraining on  the in-domain dataset $U$;
(2) both the in-domain pretraining strategy and the pseudo-label based strategy introduce 
significant performance boosts, 
with the former performing better with larger $U$,  the latter performing better with smaller $U$, and the combination leading to the largest performance gain; 
(3) vanilla self-training (pretraining first on the pseudo-label dataset $D'$ and then fine-tuning on $D$) yields better performances when $D$ is small, while joint training on the combination of 
$D'$ and $D$ yields better performances when $D$ is large. 

Using semi-supervised learning strategies, we are able to achieve a performance of around $93.8\%$ accuracy with only 50 training data points on the IMDB dataset, and 
 a competitive performance of 96.6$\%$ with the full  IMDB dataset. 
Our work marks an initial step toward understanding the behavior of semi-supervised learning models under the context of large-scale pretraining.\footnote{Code, models and datasets  can be found at https://github.com/ShannonAI/Neural-Semi-Supervised-Learning-for-Text-Classification}
\end{abstract}
\section{Introduction} 
Because of the fact that
obtaining 
 supervised training labels is costly and time-intensive, 
 and that   unlabeled data is relatively easy to obtain, 
 semi-supervised learning  \cite{10.7551/mitpress/9780262033589.001.0001,zhu2005semi}, which 
utilizes 
in-domain
 unlabeled data $U$ to improve models trained on the labeled dataset $D$, is of growing interest. 
Under the context of large-scale of language model pretraining \citep{devlin2018bert,yang2019xlnet,yinhan2019roberta,lewis2019bart,bao2020unilmv2,danqi2020spanbert},  where a language model is pretrained on an extremely large, open-domain dataset (denoted by {\it largeU}, with {\it |largeU|} $\gg$ {\it |U|}),
  how we can make the best use of the in-domain unlabeled dataset $U $
   is poorly understood. 
 There are basically two ways to take advantages of  the unlabeled, in-domain dataset $U$: 
 {\bf in-domain pretraining}\footnote{To note,  
 the pretraining on the in-domain dataset $U$ is distinguished from the pretraining on the large-scale, open-domain dataset {\it largeU}.
 The model for in-domain pretraining can be randomly initialized or 
 taking a pretrained model based on the open-domain dataset {\it largeU} \citep{dontstoppretraining2020}.}, where a language model is pretrained on 
 the in-domain dataset
 $U$, and then   fine-tuned on $D$; 
{\bf pseudo-label} based approach \citep{lee2013pseudo,reed2015training,iscen2019label,Shi_2018_ECCV,arazo2020pseudo}, where unlabeled data points are assigned with labels predicted by the model trained  on $D$ (referred to as the teacher model), forming a new dataset $D'$. 
A new model (referred to as the student model) is trained for final predictions by considering $D'$. 

Many important questions regarding the behavior of semi-supervised learning models under the context of large-scale LM pretraining 
 remain unanswered:  
 Is semi-supervised training
 still beneficial with the presence of large scale pretraining on {\it largeU}? 
  Should $U$ be used for in-domain LM pretraining or pseudo-label generation?
 How should  pseudo-label based semi-supervised models 
  be  implemented?
How different semi-supervised strategies (e.g., self learning) affect performances regarding $D$ of different sizes, and $U$ of different sizes, etc.

In this paper, we conduct comprehensive studies on the behavior of semi-supervised learning in NLP 
with the presence of large-scale language model pretraining.  
We use the task of text classification as an example,  the method of which can be easily adapted to different NLP tasks.
Our work sheds important lights on the behavior of semi-supervised learning models: 
we find that
(1)  with the presence of  in-domain pretraining LM on $U$, open-domain LM pretraining \cite{devlin2018bert} 
is unnecessary, and we
are able to achieve better performance with pretraining on  the in-domain dataset $U$;
 (2)
 both
the in-domain pretraining strategy and the pseudo-label based strategy
 lead to significant performance boosts,
with the former performing better with larger $U$, the latter performing better with smaller $U$, and the 
combination 
 of both performing the best;
(3) for pseudo-label based strategies, 
self-training (pretraining first on the pseudo-label dataset $D'$ and then fine-tuning on $D$) yields better performances when $D$ is small, while joint training on the combination of 
$D'$ and $D$ yields better performances when $D$ is large.
 
Using semi-supervised learning models, we are able to achieve a performance of around $93-94\%$ accuracy with only 50 training data points on the IMDB dataset, and 
 a competitive performance of 96.6$\%$ with the full dataset. 
More importantly, our work marks an initial step toward understanding the behavior of semi-supervised learning models in the context of large-scale pretraining.

The rest of this paper is organized as follows: related work is detailed in Section 2. 
Different strategies for training semi-supervised models are shown in Section 3. 
Experimental results and findings are shown in Section 4, followed by a brief conclusion in Section 5. 

\section{Related Work}
\subsection{Semi-Supervised Learning}
The goal of semi-supervised learning \citep{10.7551/mitpress/9780262033589.001.0001,zhu2005semi} is to use massive
amount of 
 unlabeled data to improve the models trained  on labeled data. 
One widely-used type of semi-supervised method is 
the pseudo-label  based method
 \citep{lee2013pseudo,reed2015training,iscen2019label,Shi_2018_ECCV,arazo2020pseudo}, where unlabeled data points are assigned with labels predicted by a trained model, 
forming a large pseudo labeled dataset to train a model. 
{\bf Self-training} \citep{1053799,riloff2003learning} is a specific type of pseudo-label  based method that is of growing interest. 
Self-training involves training two models: a ``teacher'' used to label unlabeled data, which is  used as an augmented labeled dataset. 
Then  a ``student'' is trained on the newly augmented dataset. 
This process can be iterated to further boost performances. Self-training has been successfully applied in 
different fields such as 
computer vision \citep{yalniz2019billion,babakhin2019semi,xie2020self,chen2020big,zoph2020rethinking}, automatic speech recognition (ASR) \cite{parthasarathi2019lessons}. 
In Vision, \citet{yalniz2019billion} adopted the self-training paradigm in image classification, and achieves the state-of-the-art top-1 result on ImageNet benchmark; \citet{xie2020self} proposed the strategy of Noisy Student Training, a variant of self-training built on top of EfficientNet \citep{tan2020efficientnet}.
In ASR, \citet{parthasarathi2019lessons} trained an ASR model on one million hours of unlabeled speech data  using a teacher-student self-training model.
\citet{park2020improved} proposed the concept of normalized filtering score that filters out low-confident utterance-transcript pairs generated by the teacher to mitigate the noise introduced by the teacher model.





In the context of natural language processing (NLP), the concept of semi-supervised learning has been adopted in different NLP tasks such as 
  machine translation \citep{cheng-etal-2016-semi,tu2016neural,ramachandran2016unsupervised,edunov2018understanding,clark2018semisupervised}, information extraction \citep{liao2009simple,peters2017semisupervised}, text classification \citep{nigam2006semi,dai2015semi,miyato2016adversarial,howard2018universal,karamanolakis2019leveraging,li2019learning}, and text generation \citep{zang2019semisupervised,qader2019semisupervised,shang2019semisupervised}.
Particularly, \citet{he2019revisiting} studied the efficacy of self-training on sequence generation tasks and  found that self-training can significantly boost performances, particularly when labeled data is scarce. Besides, they also pointed out that the core of self-training for sequence generation tasks is the noise injected into the neural model, which can be interpreted to smooth the latent sequence space.

Word vector models \citep{mikolov2013distributed,mikolov2013efficient,pennington2014glove,mikolov2013efficient,wordvec2014matrixfactor} and language modeling pretraining  \citep{devlin2018bert,yang2019xlnet,yinhan2019roberta,lewis2019bart,bao2020unilmv2,danqi2020spanbert} 
 can also be viewed as a specific type of semi-supervised learning model, by leveraging the information of data in the general domain. 
Other strategies for training semi-supervised models 
 involve co-training \citep{qiao2018deep,DBLP:conf/eccv/ChenZG18,han2018co},  low-density separation \citep{grandvalet2005semi,dai2017good}.



\subsection{Data Augmentation}
Data augmentation 
aims at 
 increasing the amount of training data by adding slightly modified copies of  existing data points or  created new synthetic data based on existing data points \citep{krizhevsky2012imagenet,paulin2014transformation,laine2016temporal,sajjadi2016regularization,cubuk2018autoaugment,inoue2018data,cubuk2020randaugment}.
 The concept consistency training \citep{rasmus2015semisupervised,sajjadi2016regularization,laine2017temporal,tarvainen2018mean,miyato2018virtual,luo2018smooth,athiwaratkun2019consistent,li2019decoupled,verma2019interpolation,liu2020decoupled} is widely used as a regulation to force the label of modified copies to the same as the original label. 
 

In NLP, 
 modified copies of  existing data points
are generated usually by 
 synonym replacement and text editing \citep{zhang2015character,kobayashi2018contextual,wei2019eda,gao-etal-2019-soft}, back-translation \citep{sennrich2016back-translation,edunov2018understanding,xie2019unsupervised}, noise injection \citep{wang-yang-2015-thats,xie2017data,xie2018noising}, mixup \citep{guo2019augmenting}, generation \citep{anabytavor2019data,wu2019conditional,kumar2020data}.
  Data augmentation has introduced significant performance boost  especially in low-resource scenarios \citep{fadaee2017data,bergmanis2017training,sahin-steedman-2018-data,xia-etal-2019-generalized,shleifer2019low,singh2019xlda}.

\section{Models}

\subsection{Notations}
We use the task of text  classification 
for illustration purposes, in which the goal is to assign a label $y$ to a given  input $x$. $x$ is a sequence of words. 
We have a given labeled set $D=\{x_i, y_i\}$, $i\in [1,N_D]$, where $N_D$ denotes the number of data points  in $D$. 
In addition, we have an  
in-domain
unlabeled dataset $U$ of size $N_{U}$, where  $N_{U} \gg N_{D}$.
The goal of semi-supervised learning is to explore how the unlabeled 
in-domain
dataset $U$ can be leveraged at the training time. 
At test time, inference remains the same as the original setup. 
\subsection{In-domain LM Pretraining}
A direct way to take  advantage of 
$U$ is to  pretrain  a BERT \citep{devlin2018bert}, RoBERTa \citep{yinhan2019roberta} or GPT3 \citep{brown2020language} style language model
on $U$
 by predicting a masked word given surrounding contexts or 
 predicting
 a subsequent word given proceeding contexts. 
 Pretraining on the in-domain data  
facilitates the learning of in-domain   
semantics and word compositions. 

For training, the LM model can be trained from scratch with random initialization or 
 initialized using an existing BERT or RoBERTa model trained on an open-domain dataset, the latter of which is similar to the idea in \newcite{dontstoppretraining2020}, which continues LM pretraining in different domains and tasks. 
 The LM model trained on $U$ is used as initialization to be further finetuned on $D$.

\subsection{Pseudo-label Based Approach}
Another way to take advantage of $U$ is to use the model trained on $D$ to assign
pseudo
 labels to data points in $U$. 
Specifically, 
the model trained on $D$ is referred to as the {\it teacher} model. 
The teacher model is used to 
  assign pseudo labels to $U$ or a specific portion  of $U$, forming  the augmented  dataset $D'$. 
Let $N_{D'}$ denote the number of selected examples in  $D'$.
There are  different options on how to generate $D'$,  how 
 $D$ and $D'$ are combined, and how the new model can be trained on the  combination. 
The model trained on the combination is referred to as the {\it student} model. 

The teacher  and the student can share a similar model backbone. 
In the  text classification task, various structures can be used as the backbone such as LSTMs \citep{hochreiter1997long,tang-etal-2016-effective}, 
CNNs \citep{kim-2014-convolutional}, BERT \cite{devlin2018bert,chai2020description}, etc, with the trained model to maximize the probability of 
predicting
the golden label $y$ given $x$. 
The {\it student} model can be randomly initialized or initialized using the {\it teacher} model.

\subsubsection{Different strategies to construct $D'$}
Here we discuss different strategies to generate $D'$.
\paragraph{Naive Strategy $D' = U$:}
The teacher model is used to label all instances in $U$ to form $D'$. 
The  shortcoming for this strategy is that incorrect and low-confident labels 
  included in 
 $D'$ can be severely detrimental to the student model. 

\paragraph{Top-K Model Predictions:} To avoid the negative effect from the 
low-confident 
 labels assigned by the teacher, we can only pick the confident ones. 
The teacher  model  is run on each example in $U$ to obtain the probability of all classes. 
Then for each class $l$, we rank all instances in $U$ based on the corresponding probabilities. The top-$K$ instances for each class are selected to form  $D'$. 

\subsubsection{Different Strategies for Student Training }
Here we discuss how models can be trained given $D'$ and $D$. 
\paragraph{Training on $D+ D'$:}
This strategy is denoted by $T(D+D')$.
The most straightforward strategy is to train the student model on the union of $D$ and $D'$. 
The potential risk with this naive strategy is that 
the influence from clean labels in $D$ can be diluted if the size of $D'$ is large and that 
 incorrect labels in $D'$ can exert  negative effects on 
the student model.

\paragraph{Pretraining on $D'$ and Fine-tuning on $D$:}
This strategy is denoted by $T(D')F(D)$.
To make the model more immune to incorrectly labeled examples in $D'$, and be able to take advantage of the large amount of data in $D'$ at the same time, we can first 
train the student model
 on the newly collected dataset $D'$ to predict the pseudo labels, and then fine-tune the model on the original labeled dataset $D$. 
This strategy 
 ensures that
the final model is fine-tuned on the dataset with clean labels.
This process actually mimics the idea of the self-training \citep{devlin2018bert,xie2020self,chen2020big,grill2020bootstrap} in semi-supervised learning literature. 

\paragraph{Pretraining on $D+D'$ and Fine-tuning on $D$:}
This strategy is denoted by $T(D+D')F(D)$. 
A minor change can be made to the strategy described above, where the model is pretrained on the concatenation of $D$ and $D'$, and then fine-tuned on $D$. 
Practically, 
we find that  $T(D+D')F(D)$ consistently outperforms 
$T(D')F(D)$, which is expected since adding examples with golden labels for pretraining does no harm the performance. 
We thus only report results for $T(D+D')F(D)$, and omit $T(D')F(D)$ for brevity. 

\paragraph{Iterative Training}
The process of training the teacher and the student 
 can be iterated: 
the student model trained in the previous iteration 
 can be used 
as  a new teacher model  
to   relabel the unlabeled dataset $U$, 
       from which the top-$K$ examples are regenerated to form the new $D'$. 
       Based on the new $D'$, a new student model is trained. 
       This process is repeated until the pre-defined value of iterations $N$ is reached.

\subsection{Combining In-domain LM Pretraining and Pseudo-label Based Approach}
The in-domain LM pretraining and the pseudo-label based strategy can be combined: 
an LM model is first pretrained on the in-domain data $U$. Next, the model is used  to initialize the teacher, which will be trained on $D$. 
The teacher model is then used to generate pseudo labels for $U$, which are used to train the student.
In this way, $U$ is used twice, both in LM pretraining and pseudo-label generation for student training.

\section{Experiments}
In this section, we conduct extensive experiments to better understand the behavior of semi-supervised learning. 
We give insights gathered from extensive experimental studies and also discuss several considerations towards obtaining  a successful model.
We use the following two  datasets for detailed explorations: 

(1) The labeled dataset $D$ is the
 IMDB dataset collected by \citet{maas2011learning}. 
This dataset contains an even number of positive and
negative movie reviews. The training and test sets respectively contain 25k and 25k examples.
The task is formalized as a binary classification task to decide the polarity of sentiment for a review. 
To explore models' behavior on training datasets of different sizes, we use 10, 20, 50, 100, 1k, 5k and 25k for training. 

For the unlabeled dataset $U$, we crawled the IMDB and collected about 3.4M movie reviews.
We release this large-scale IMDB movie review dataset  to public.

(2) The labeled dataset $D$ is the deceptive  opinion spam dataset \citep{ott-etal-2011-finding,li-etal-2014-towards} to 
separate fake hotel reviews generated by Turkers and hotel employees from genuine reviews from real customers.  
The task is formulated as a three-class classification task, and  
$D$ contains 800/280/800 reviews, which respectively denote fake reviews from Turkers, fake reviews from hotel employees and genuine reviews from real customers.
For the unlabeled dataset $U$, we crawled TripAdvisor, an online travel  website and collected about 1M reviews from roughly 5k hotels. 

The sentiment analysis task on movie reviews is a significantly easier task than the deceptive review detection task, 
 since the latter requires to model to identify subtle changes in language usage for spam generation, while the former focuses more on identifying sentiment-indicative tokens. 

\paragraph{Training Details}
For in-domain LM pretraining, we use the 
RoBERTa structure \citep{yinhan2019roberta} as the backbone. 
We use both the small model, a 12-layer transformer with the hidden size of each layer being 768,
and a large model, a 24-layer transformer with the hidden size of each layer being 1,024.
Models are trained using using Adam \citep{kingma2014adam} with $\beta=(0.9,0.98)$, $\epsilon=10^{-6}$, a polynomial learning rate schedule, warmup for 4K steps and weight decay with $10^{-3}$. Dropout rate is set to 0.2. 
The pretrained LM model on the in-domain $U$ is then finetuned on $D$.
For training the teacher and the student,  we use Adam for optimization.  Learning rate, batch-size and dropout rate are treated hyperparamters tuned on the dev set. 

For reference purposes,  we also implement the BiLSTM model, where the representation at the last time step from the left-to-right direction is concatenated with the 
the representation at the first time step from the right-to-left direction, which is next fed to the softmax function for golden label prediction. 
Word vectors are initialized using GloVe  \citep{pennington2014glove}. 

\begin{table*}[t]
\renewcommand\arraystretch{1.2}
\centering
\small
\setlength{\tabcolsep}{1mm}{
\begin{tabular}{c|c|cccc|cccc|cccc}
  \toprule
  & T$(D)$ & \multicolumn{4}{c}{T$(D')$} & \multicolumn{4}{c}{T$(D+D')$} & \multicolumn{4}{c}{$ \text{T}(D+D')\text{F}(D) $} \\\hline
  \diagbox[width=1.6cm,height=0.82cm]{$|D|$}{$|D'|$} & 0 & 1K & 10K & 100K & 1M & 1K & 10K & 100K & 1M & 1K & 10K & 100K & 1M \\\hline
  10 & 53.34 & 54.00 & 51.46 & 53.95 & 53.60 & 55.10 & 51.54 & 52.49 & 53.91 & 55.25 & 55.35 & {\bf 55.47} & 54.60\\
  && (+0.66)  &  (-1.88) & (+0.61)  &  (+0.26) & (+1.76)  &  (-1.80) &  (-0.85) & (+0.57)  &  (+1.91) &  (+2.01) & {\bf (+2.13)}  &(+1.26)\\\hline
  20 & 54.62 &  51.73 & 53.28  & 54.16  & 55.43  & 54.35  & 54.50  &  53.92 & 54.43  & { 58.15}  & 58.63  &{\bf 58.77}  & 55.63\\
  && (-2.89)  &  (-1.34) & (-0.46)  &  (+0.81) & (-0.27)  &  (-0.12) &  (-0.70) & (-0.19)  &  { (+3.53)} &  (+4.01) &  {\bf (+4.15)} &(+1.01)\\\hline
  50 & 80.67 & 56.91  & 72.25  & 82.19  &  81.30 & 73.81  & 77.36  & 83.80  & 81.63  & 81.32  &  84.96 & {\bf 85.86}  & 82.85\\
  && (-23.76)  &  (-8.42) & (+1.52)  &  (+0.63) &(-6.86)  &  (-3.31) &  (+3.13) & (+0.96)  & (+0.65) &  (+4.29) &  {\bf (+5.19)} &(+2.18)\\\hline
  100 & 83.12 &  65.67 &  83.49 & 85.52  & 83.95  & 80.30  & 83.50  & 84.89  & 83.60  & 81.86  & 86.25  & {\bf 87.27}  & 84.40\\
  && (-17.48)  &  (+0.37) & (+2.40)  &  (+0.83) & (-2.82)  &  (+0.38) &  (+1.77) & (+0.48)  &  (-1.26) &  (+3.13) &{\bf (+4.15)}  &(+1.28)\\\hline
  1K & 89.90  &  -- &  84.68 & 89.84  & 90.66  & --  & 90.10  &  90.15 & 91.12  &  -- & 90.45  & 90.78  &{\bf 91.32} \\
  && -- &  (-5.22) & (-0.06)  &  (+0.76) &--  &  (+0.20) &  (+0.25) & (+1.22)  & -- &  (+0.55) & (+0.88)  &{\bf (+1.42)}\\\hline
  5K & 93.16 & --  & 87.09  & 91.43  & 92.98  & --  & 93.02  & 93.57  &  {\bf 93.79} & --  & 93.07  & 93.05  & 93.72\\
  && --  &  (-6.07) & (-1.73)  &  (-0.18) & --  &  (-0.14) &  (+0.41) & {\bf (+0.63)}  & -- &  (-0.09) &  (-0.11) &(+0.56)\\\hline
  25K & 95.20 & --  & --  & 90.56  & 92.75  & --& --  & 95.51  & {\bf 95.80}  &  -- & --  &  95.42 &{ 95.50} \\
  && --  &  -- & (-3.24)  &  (-1.05) & --  &  -- &  (+0.31) & (+{\bf 0.60})  &  -- &  -- &  (+0.22) &{ (+0.30)}\\
  \bottomrule
\end{tabular}
}
\caption{Results for different  strategies for
pseudo-label based approaches
 on the IMDB dataset based on {\bf open-domain pretraining}. ``T'' represents ``Train'' and ``F'' represents ``Fintune''. Brackets represent the used dataset(s), e.g., ``T(D+D')F(D)'' means the model is trained on $D+D'$ and finetuned on $D$. The best result for each row is marked in bold.}
\label{tab:results}
\end{table*}
\begin{figure}[t]
  \centering
  \includegraphics[scale=0.43]{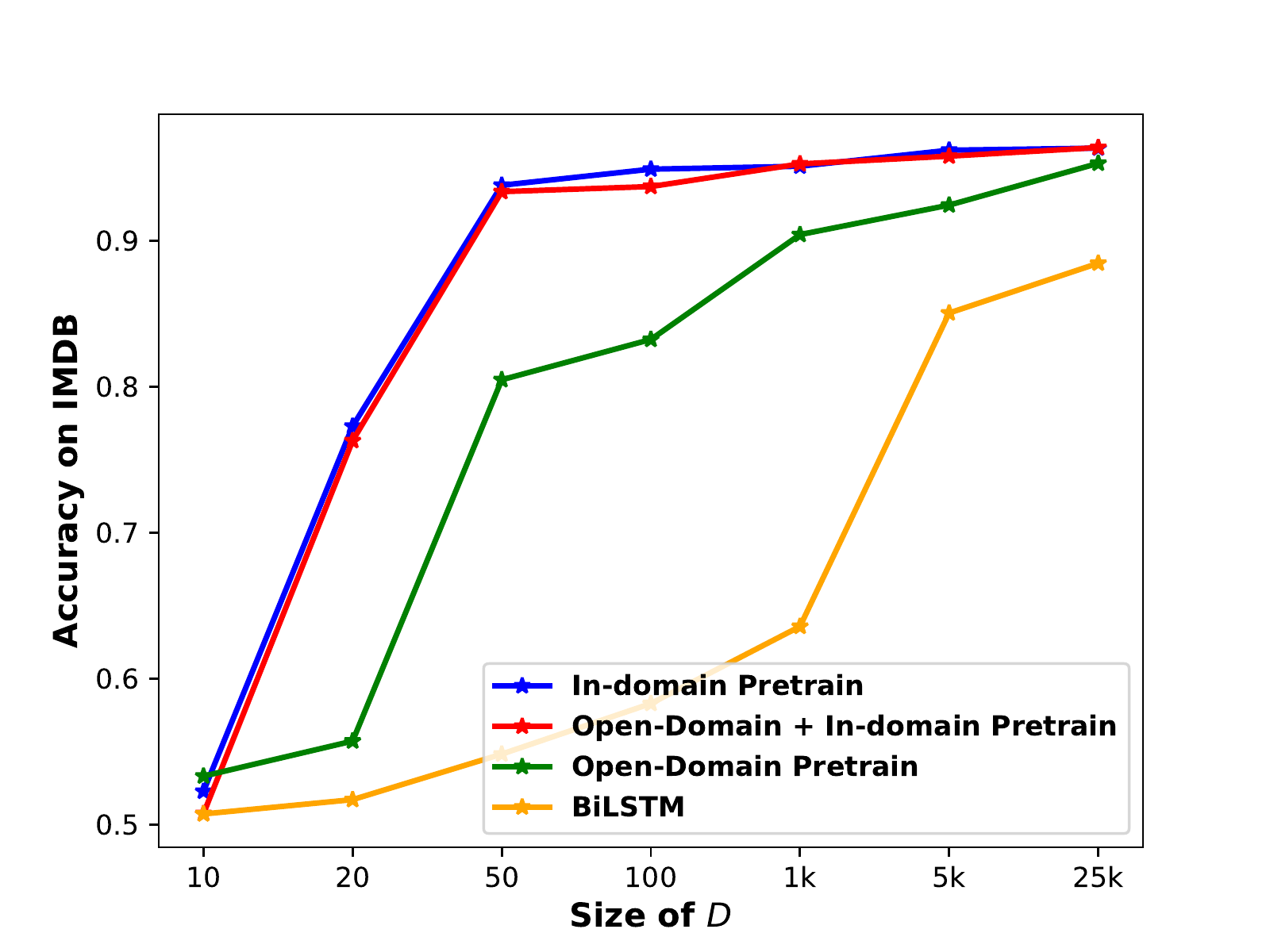}
    \includegraphics[scale=0.43]{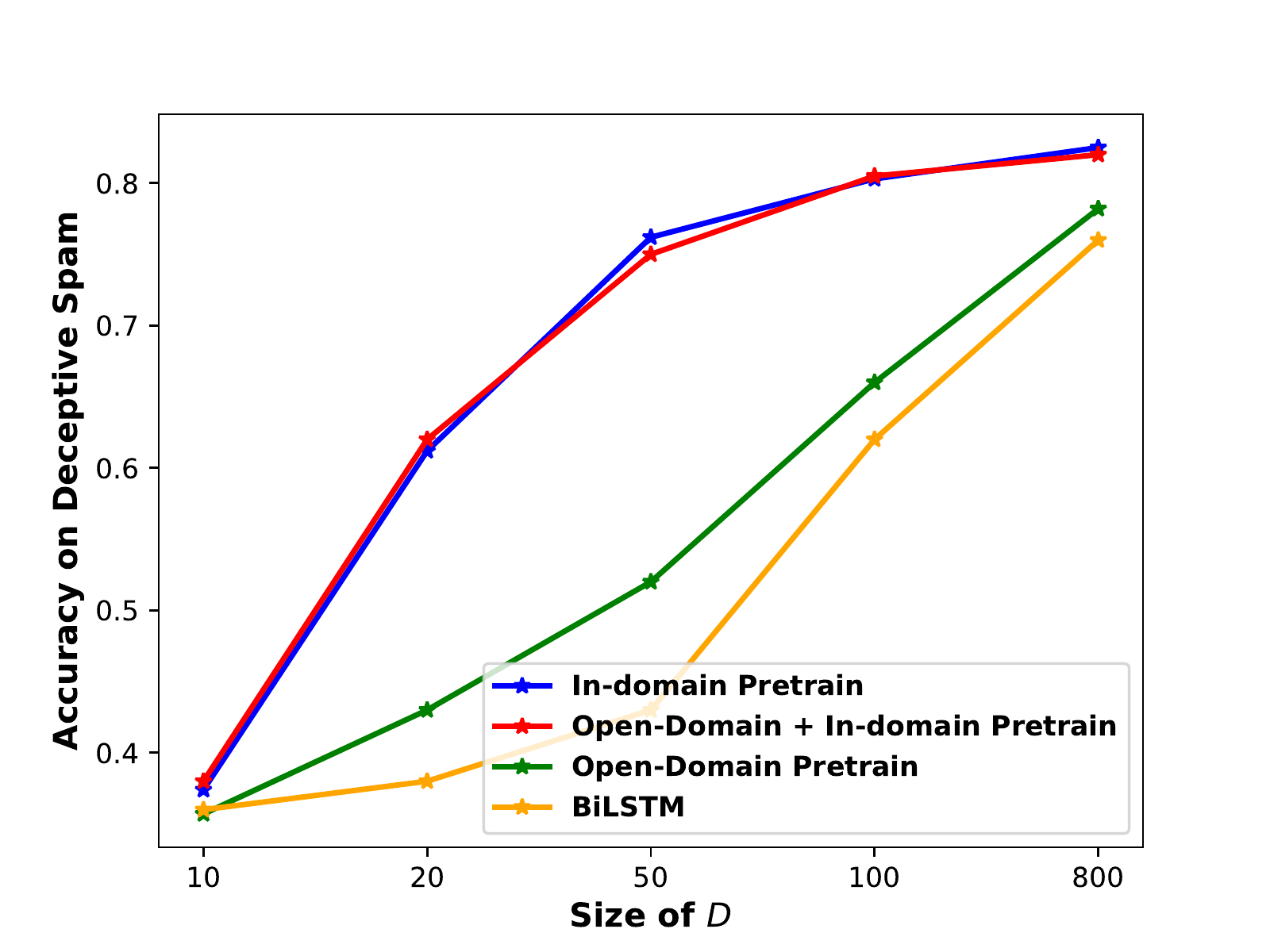}
  \caption{Teacher perforamnces for BiLSTMs,   In-domain pretraining, 
  Open-domain pretraining, 
  Open-domain+In-domain pretraining  on the IMDB (upper) and the deceptive opinion spam (lower) datasets based on different sizes of the labeled dataset. All pretraining models use RoBERTa small as the backbone.  }
  \label{fig:general}
\end{figure}

\subsection{Teacher Performances}
We first examine  teacher performances
 for BiLSTMs,  {\it In-domain pretraining}, {\it Open-domain pretraining} and {\it Open-domain+In-domain  pretraining} 
on the IMDB and the deceptive  spam  datasets 
regarding different values of $|D|$, as 
 shown in Figure \ref{fig:general}. 
For {\it In-domain pretraining}, a 
randomly initialized 
LM model is  first pretrained on the in-domain dataset $U$, and then fine-tuned  on  $D$.
For {\it Open-domain pretraining},
we  directly take the pretrained RoBERTa model, which is  pretrained on the open-domain dataset $largeU$. 
For   {\it open-domain+In-domain  pretraining}, the LM model is first pretrained on  $largeU$, then on the in-domain dataset $U$, and last fine-tuned  on $D$. 
The pseudo-label based semi-supervised strategy is  applied in none of these setups. 
The accuracy progressively increases 
as we increase the size of the labeled dataset $D$, which is in line with our expectation.

\paragraph{Is open-domain  pretraining still necessary?}
Take the IMDB dataset as an example. 
As can be seen from Figure \ref{fig:general}, 
the performance of {\it In-domain pretraining} 
(95.87 when $|D|=25k$)
on the 3.4M unlabeled reviews performs 
nearly 
 the same as {\it open-domain+In-domain  pretraining} (95.82), both of which significantly outperform 
{\it open-domain pretraining} (95.20). 
Similar phenomena are observed for the deceptive spam dataset, where
{\it In-domain pretraining},  {\it open-domain+In-domain  pretraining} and {\it open-domain pretraining} respectively 
obtains 82.5, 82.0 and 78.2 accuracy. 
This demonstrates that with the presence of relatively large in-domain data, the extremely time-intensive training of
 LM on a huge amount of open-domain data is unnecessary. 

\subsection{In-domain Pretraining}
\begin{figure*}[!ht]
  \includegraphics[scale=0.38]{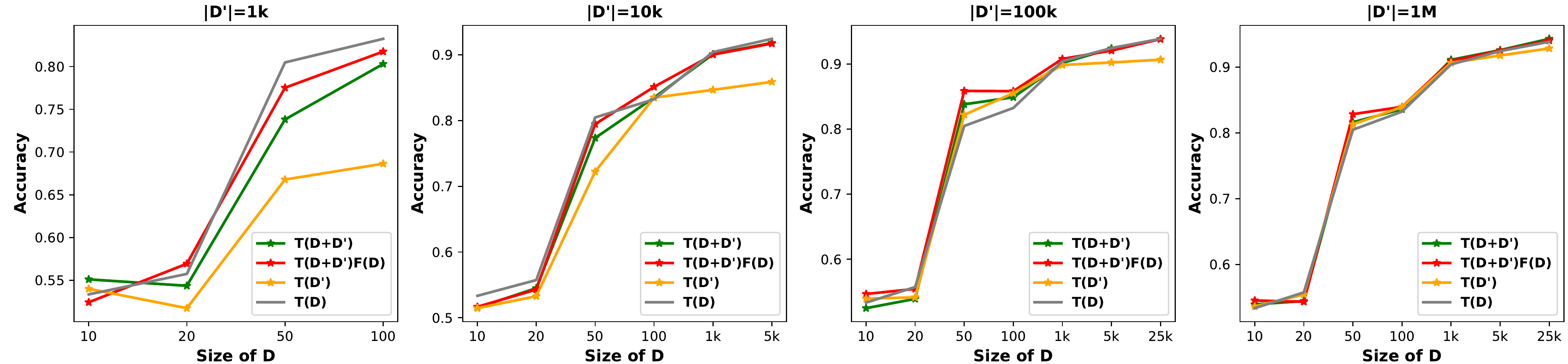}
  \caption{The influence of $|D|$ on pseudo-label based approaches for different $|D'|$.  }
  \label{fig:size_d}
\end{figure*}

As shown in Figures \ref{fig:general} and \ref{fig:size_d}, 
the  in-domain pretraining strategy has significantly better {\bf few-shot} learning abilities and
requires much smaller amount of  data for training:
the performance 
from in-domain pretraining
drastically improves as  $|D|$ increases from 10 to  50, 
achieving a performance of 93.8 accuracy on the IMDB dataset with only 50 training examples.
This performance is 
higher 
 than BiLSTMs with 25K training examples, and similar to the performance of vanilla RoBERTa with 5K training examples. 
This shows that  LM pretraining on the in-domain dataset $U$ provides the model with the generality to rapidly figure out the necessary task-specific information for predictions. 

Comparing with existing  pretraining models that have few-shot learning ability such as GPT3, the advantages of the semi-supervised in-domain pretraining are obvious: the model
is easier to train, 
 has significantly fewer parameters and does not have to rely on a vast amount of training data.

\subsubsection{Influence from the size of $|U|$}
\begin{figure}[t]
\center
    \includegraphics[scale=0.45]{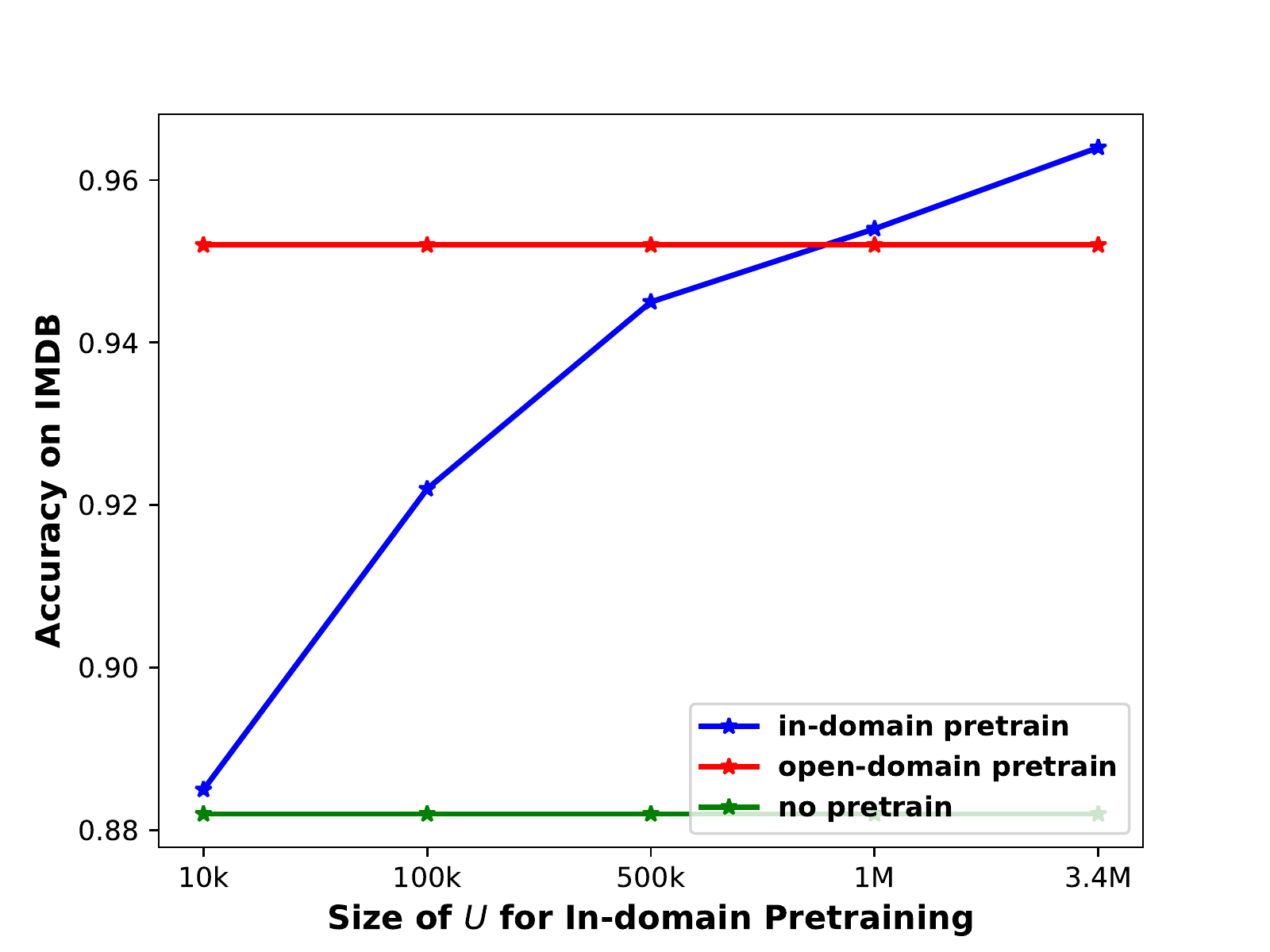}
  \caption{The influence of $|U|$ for in-domain pretraining with $|D|=25k$.  }
  \label{fig-U-pretrain}
\end{figure}

It is widely accepted that LM pretraining requires a massive amount of training data. 
Results for $U$ 
of different sizes 
for LM pretraining are shown in  Figure \ref{fig-U-pretrain}.
The model is randomly initialized and then trained on the in-domain $U$ until convergence. 
The pretrained LM model is next 
 fine-tuned on $D=25k$. 

As can be seen, the performance of {\it in-domain pretraining} highly relies on the size of the in-domain dataset $U$. 
With smaller sizes of $U$, {\it in-domain pretraining}  underperforms {\it open-domain pretraining}, which is in line with our expectation since small $U$  cannot provide enough evidence for learning in-domain word semantics and compositions.
Specifically, the performance of $|U|=10k$ is only slightly better than the {\it no-pretraining} setup;
the performance  for {\it in-domain pretraining} outperforms {\it open-domain pretraining} only when the size of $U$ exceeds 1M. 

\begin{figure}[t]
\center
  \includegraphics[scale=0.45]{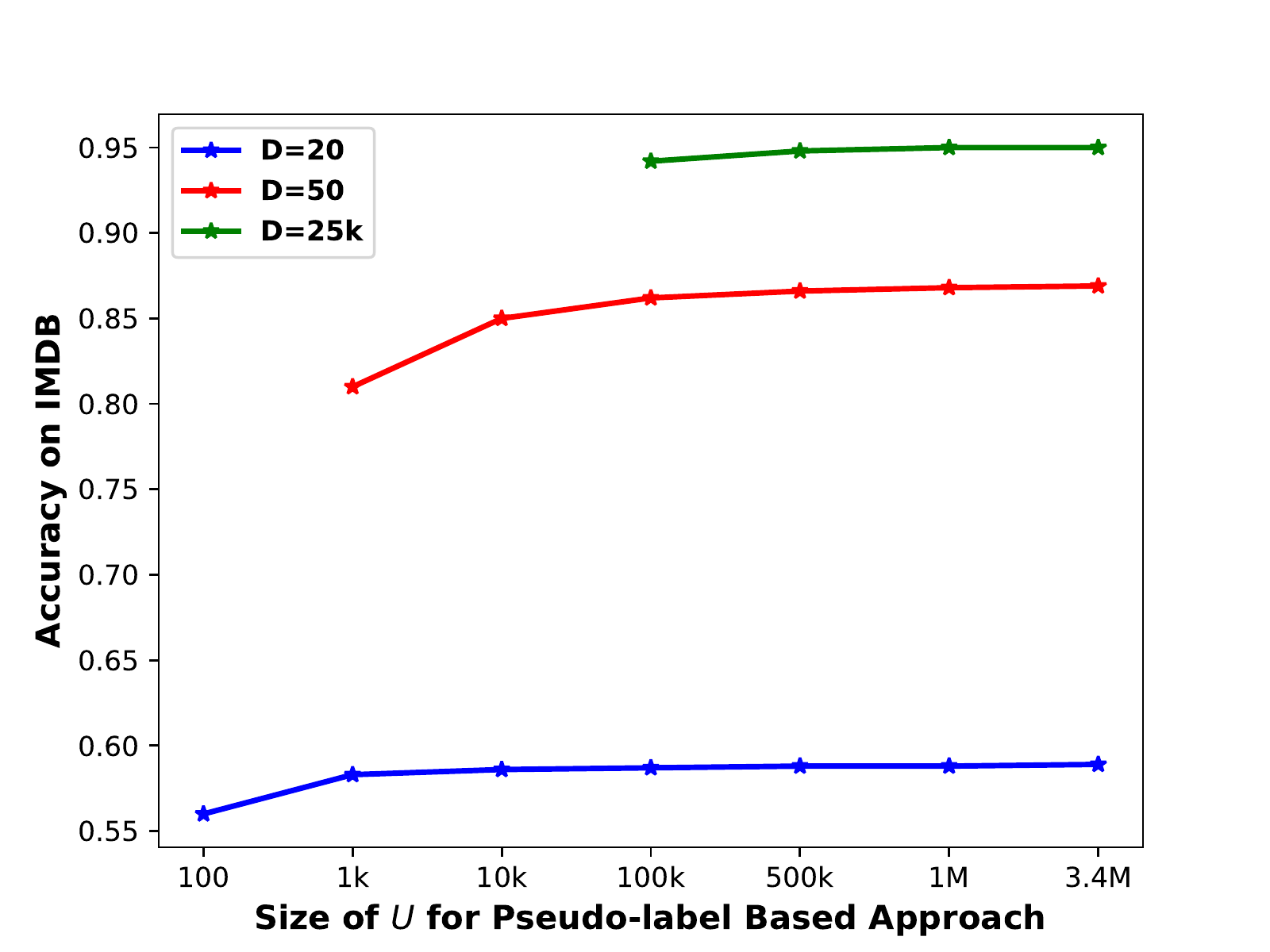}
  \caption{The influence of $|U|$  for pseudo-label based approaches.  }
  \label{fig-U-pseudo}
\end{figure}

\begin{figure*}[!ht]
  \includegraphics[scale=0.38]{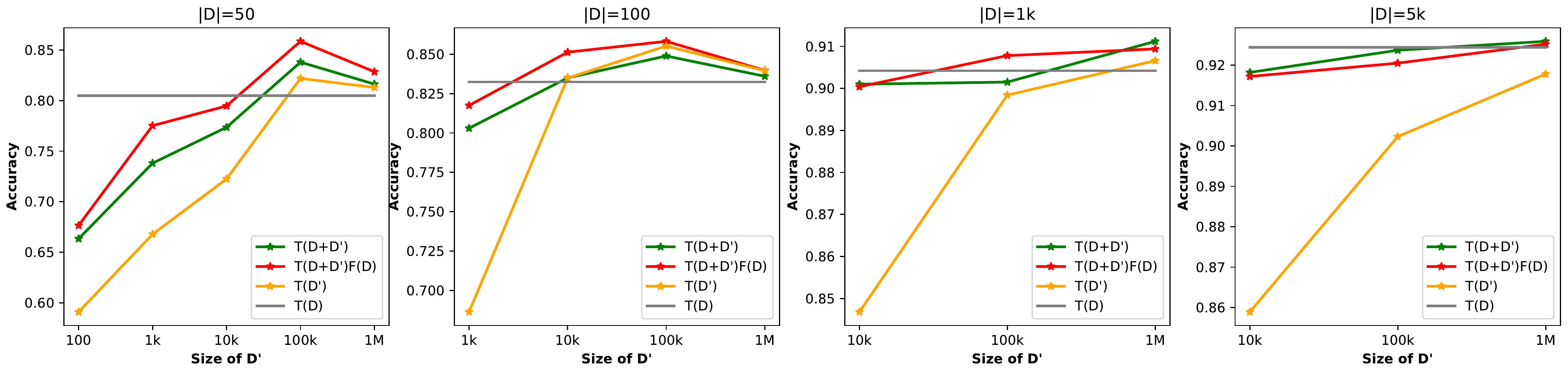}
  \caption{The influence of $|D'|$ on pseudo-label based approaches. $|D|$ is required to be larger than $|D|$. }
  \label{fig:size_d_}
\end{figure*}

\subsection{Pseudo-label Based Approaches}
Detailed results for 
pseudo-label based approaches 
based on open-domain pretraining
are shown in Table \ref{tab:results}.
The  trends can be summarized as follows: 

(1) Generally, both $\text{T}(D+D')\text{F}(D)$ and $\text{T}(D+D')$ 
perform better than  the vanilla setup $T(D)$.
$T(D')$ underperforms the vanilla setup $T(D)$ when $D$ is small and $D$ is large, 
but outperforms $T(D)$ when $D$ is of medium size. 

(2) The performance boost 
introduced by pseudo-label based methods gradually increases as the size of $D$ increases, and then shrinks.
The explanation is as follows:
with a small $D$, the accuracy of the teacher model is low.
Most predicted labels on $D'$ are  thus unreliable. 
Therefore, 
the advantage from the model trained on the noisy $D'$ is relatively small;
with a large $D$, the teacher model is already good enough, reaching an accuracy higher than 0.9. 
Though
almost all 
 predicted labels on $D'$ are  correct, their improvement upon an already pretty good teacher model is small.
 There is a sweet spot for the size of $D$, where  pseudo-label based methods introduce the largest boost: 
 with a medium-sized $D$, where the teacher model reaches an acceptable accuracy when the correctly labeled examples in $D'$
outweigh incorrectly labeled ones, the student model can take the most advantage of $D'$, leading to significant performance boosts of +5$\%$. 
Similar trends are observed for $T(D')$, where $T(D')$ underperforms the vanilla setup $T(D)$ both when $D$ is small and $D$ is large, 
but outperforms $T(D)$ when $D$ is of medium size.

(3) For the three pseudo-label based strategies, $T(D')$, $\text{T}(D+D')\text{F}(D)$ and $\text{T}(D+D')$, 
$T(D')$ performs the worst for all $|D|$.  This is in line with our expectation because for $T(D')$,  only $D'$ is used for final predictions, and thus the influence of the golden-labeled dataset $D$ is diluted;
$\text{T}(D+D')\text{F}(D)$ works best  
when $|D|$ is small, while 
$\text{T}(D+D')$ works best when $|D|$ is large.
Our explanations are as follows: 
when $|D|$ is small,  the student trained on $D+D'$ is inferior due to the massive amount of incorrect labels in $D'$. 
The model thus needs to be further fine-tuned on $D$, making $\text{T}(D+D')\text{F}(D)$ 
perform better than $\text{T}(D+D')$; 
When $|D|$ is large, most labels in $D'$ are correct and
the student is more immune to the small proportion of incorrect labels in $D'$. 
Directly training on the larger $D+D'$ dataset provides the model with more generalization ability,
while fine-tuning only on $D$ dilutes the influence from massive amount of correct labels in $D'$, making the performance of 
$\text{T}(D+D')\text{F}(D)$  worse than $\text{T}(D+D')$ when $|D|$ is large. 

\subsubsection{Influence from the size of $D'$}
With a fixed in-domain dataset $U$, we select top-K examples for each label $l$.
Different values of $K$ lead to different sizes of $D'$. 
There is apparently a tradeoff between the size of $D'$ and the confidence for examples included in $D'$:
larger size of $D'$ means that more less-confident examples are selected.

Trends regarding different values of $|D'|$
are shown in Figure \ref{fig:size_d_}.
We only run setups with $|D'|$  larger than $|D|$. 
As can be seen, for smaller $D$ ($|D|= 50, 100$), the final performance first increases as $|D'|$ gets larger, which means the model is taking  advantage of evidence provided by the pseudo labels. Then, the performance decreases as $|D'|$ continues to increase, which means the model starts suffering from the incorrectness of the less-confident examples. 
For larger values of $D$ ($|D|$ larger than 1k),  performances keep increasing as $|D'|$ gets larger. This is because the teacher model trained on $D$ is good enough to generate confident examples.

\subsubsection{Influence from the size of $U$}
The influence of the size of $U$ is 
already 
 manifested in the size of $D'$, since the size of $U$ should be larger than the size of $D'$ ($D'$ is selected from $U$). 
 Performances for different sizes of $U$ are shown in Figure \ref{fig-U-pseudo}, where we randomly sample examples from the 3.4 million IMDB reviews to form $U$ of different sizes. 
 For a given $U$, we plot the performance achieved with the best $|D'|$. 
 Larger $U$ should lead to better performances since more confident examples can be selected.
 
 As shown in Figure \ref{fig-U-pseudo},
for $D$ of different sizes,
the performance first increases as $|U|$ grows, but then immediately  plateaus. 
Explanations are as follows: 
for larger $|D|$, the model is already good enough to provide confidently correct labels for points in $D'$, and the improvement from 
extra confident  examples is marginal. 

\subsubsection{Iterative Training}
\begin{figure}[t]
\center
    \includegraphics[scale=0.45]{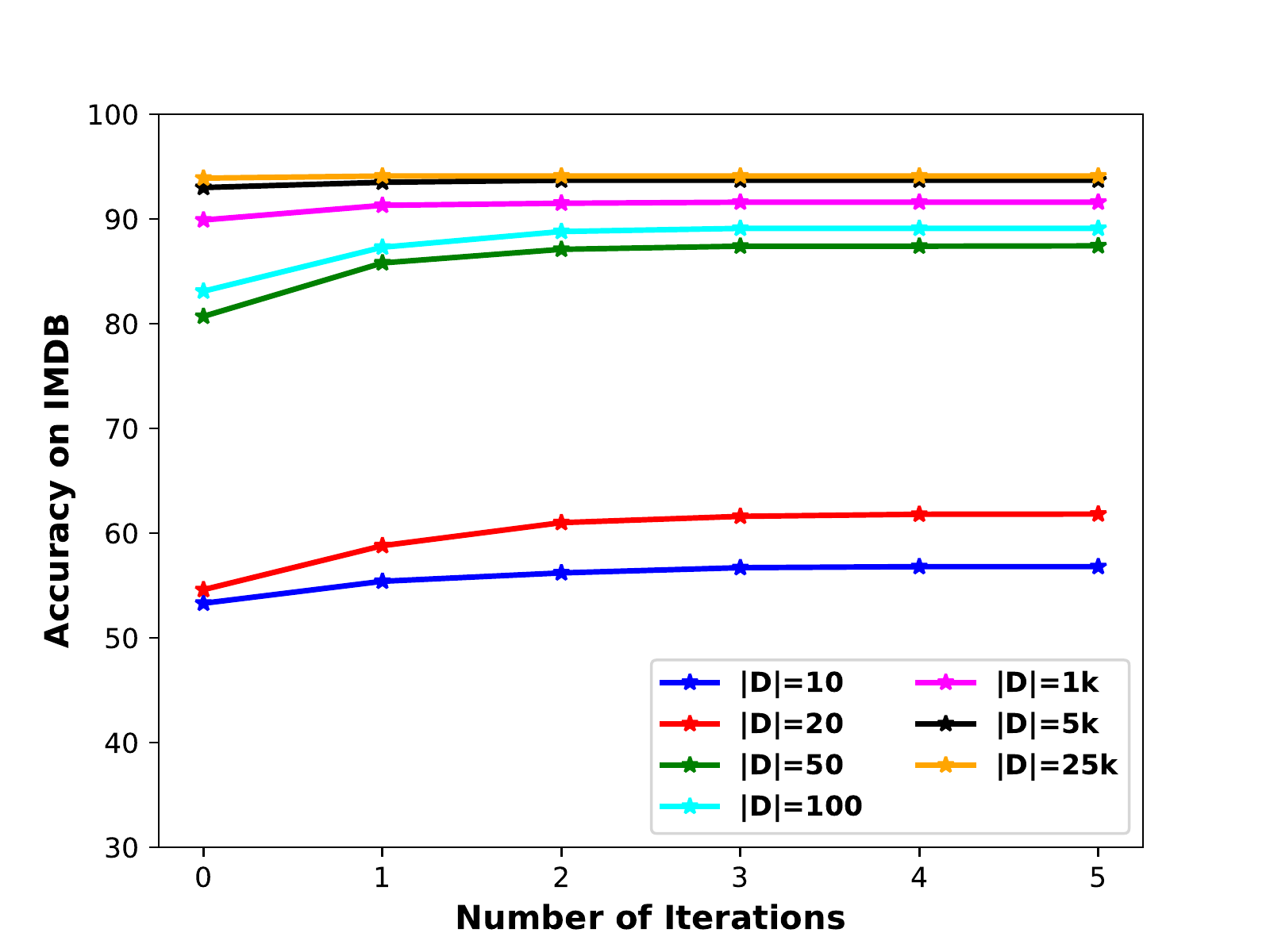}
  \caption{Iterative Training for Pseudo-label Based Approaches on IMDB based on Open-domain Pretraining.}
  \label{Iterative}
\end{figure}

We can iterate the teacher-student pattern 
for pseudo-label based approaches, where the teacher for the current iteration is initialized with the student in the previous iteration.
Results are shown in Figure \ref{Iterative}. As can be seen, additional performance boosts are observed for different $|D|$ as the iterative process goes on.
For larger $|D|$, the performance becomes stagnant across the iterative process, while for smaller $|D|$, the curve keeps rising until convergence.  

\subsection{Combing In-domain Pretraining and Pseudo-label Based Approaches}
In-domain pretraining and pseudo-label based approaches can be combined, where the teacher model is initialized with the pretrained model on the in-domain dataset $U$. 
Results
for the combined strategy 
 are shown in Table \ref{combined}.
As can be seen, both strategies lead to progressive performance boosts over Open-domain Pretraining, with a combined boost of +1.1 for the small model, and
+1.0 for the large model. 
 
\begin{table}
\center
\small
\begin{tabular}{ll}\hline
\multicolumn{2}{c}{Small |D|=25k}   \\
Open-domain Pretraining & 95.2 \\
Open-domain+In-domain Pretraining & 95.8 \\
In-domain Pretraining &95.8 \\
In-domain Pretraining + Pseudo-label & 96.3\\\hline
\multicolumn{2}{c}{Large |D|=25k}   \\
Open-domain Pretraining & 95.6 \\
In-domain Pretraining & 96.2 \\
In-domain Pretraining + Pseudo-label & 96.6\\\hline
\end{tabular}
\caption{Results for Combining In-domain Pretraining and Pseudo-label based Approaches}
\label{combined}
\end{table}

\section{Conclusion}
In this paper, we conduct comprehensive 
analysis on semi-supervised learning in NLP under the context of 
large-scale
language model pretraining. 
We find that even with the presence of large-scale LM pretraining,
both the in-domain pretraining strategy and the pseudo-label based strategy introduce 
additional 
significant performance boost, 
with the former performing better with larger $U$,  the latter performing better with smaller $U$, and the combination leading to the best performance. 
Using semi-supervised learning models, we are able to achieve a performance of around $93-94\%$ accuracy with only 50 training data points on the IMDB dataset, and 
 a competitive performance of 96.6$\%$ with the full dataset. 
Our work 
sheds light on 
 the behavior of semi-supervised learning models in the context of large-scale pretraining.

\bibliography{emnlp2020}

\begin{thebibliography}{95}
\expandafter\ifx\csname natexlab\endcsname\relax\def\natexlab#1{#1}\fi

\bibitem[{Anaby-Tavor et~al.(2019)Anaby-Tavor, Carmeli, Goldbraich, Kantor,
  Kour, Shlomov, Tepper, and Zwerdling}]{anabytavor2019data}
Ateret Anaby-Tavor, Boaz Carmeli, Esther Goldbraich, Amir Kantor, George Kour,
  Segev Shlomov, Naama Tepper, and Naama Zwerdling. 2019.
\newblock Not enough data? deep learning to the rescue!

\bibitem[{Arazo et~al.(2020)Arazo, Ortego, Albert, O’Connor, and
  McGuinness}]{arazo2020pseudo}
Eric Arazo, Diego Ortego, Paul Albert, Noel~E O’Connor, and Kevin McGuinness.
  2020.
\newblock Pseudo-labeling and confirmation bias in deep semi-supervised
  learning.
\newblock In \emph{2020 International Joint Conference on Neural Networks
  (IJCNN)}, pages 1--8. IEEE.

\bibitem[{Athiwaratkun et~al.(2019)Athiwaratkun, Finzi, Izmailov, and
  Wilson}]{athiwaratkun2019consistent}
Ben Athiwaratkun, Marc Finzi, Pavel Izmailov, and Andrew~Gordon Wilson. 2019.
\newblock There are many consistent explanations of unlabeled data: Why you
  should average.

\bibitem[{Babakhin et~al.(2019)Babakhin, Sanakoyeu, and
  Kitamura}]{babakhin2019semi}
Yauhen Babakhin, Artsiom Sanakoyeu, and Hirotoshi Kitamura. 2019.
\newblock Semi-supervised segmentation of salt bodies in seismic images using
  an ensemble of convolutional neural networks.
\newblock In \emph{German Conference on Pattern Recognition}, pages 218--231.
  Springer.

\bibitem[{Bao et~al.(2020)Bao, Dong, Wei, Wang, Yang, Liu, Wang, Piao, Gao,
  Zhou, and Hon}]{bao2020unilmv2}
Hangbo Bao, Li~Dong, Furu Wei, Wenhui Wang, Nan Yang, Xiaodong Liu, Yu~Wang,
  Songhao Piao, Jianfeng Gao, Ming Zhou, and Hsiao-Wuen Hon. 2020.
\newblock Unilmv2: Pseudo-masked language models for unified language model
  pre-training.

\bibitem[{Bergmanis et~al.(2017)Bergmanis, Kann, Sch{\"u}tze, and
  Goldwater}]{bergmanis2017training}
Toms Bergmanis, Katharina Kann, Hinrich Sch{\"u}tze, and Sharon Goldwater.
  2017.
\newblock Training data augmentation for low-resource morphological inflection.
\newblock In \emph{Proceedings of the CoNLL SIGMORPHON 2017 Shared Task:
  Universal Morphological Reinflection}, pages 31--39.

\bibitem[{Brown et~al.(2020)Brown, Mann, Ryder, Subbiah, Kaplan, Dhariwal,
  Neelakantan, Shyam, Sastry, Askell et~al.}]{brown2020language}
Tom~B Brown, Benjamin Mann, Nick Ryder, Melanie Subbiah, Jared Kaplan, Prafulla
  Dhariwal, Arvind Neelakantan, Pranav Shyam, Girish Sastry, Amanda Askell,
  et~al. 2020.
\newblock Language models are few-shot learners.
\newblock \emph{arXiv preprint arXiv:2005.14165}.

\bibitem[{Chai et~al.(2020)Chai, Wu, Han, Wu, and Li}]{chai2020description}
Duo Chai, Wei Wu, Qinghong Han, Fei Wu, and Jiwei Li. 2020.
\newblock Description based text classification with reinforcement learning.

\bibitem[{Chapelle et~al.(2006)Chapelle, Schölkopf, and
  Zien}]{10.7551/mitpress/9780262033589.001.0001}
Olivier Chapelle, Bernhard Schölkopf, and Alexander Zien. 2006.
\newblock \emph{{Semi-Supervised Learning}}.
\newblock The MIT Press.

\bibitem[{Chen et~al.(2020)Chen, Kornblith, Swersky, Norouzi, and
  Hinton}]{chen2020big}
Ting Chen, Simon Kornblith, Kevin Swersky, Mohammad Norouzi, and Geoffrey
  Hinton. 2020.
\newblock Big self-supervised models are strong semi-supervised learners.
\newblock \emph{arXiv preprint arXiv:2006.10029}.

\bibitem[{Chen et~al.(2018)Chen, Zhu, and Gong}]{DBLP:conf/eccv/ChenZG18}
Yanbei Chen, Xiatian Zhu, and Shaogang Gong. 2018.
\newblock Semi-supervised deep learning with memory.
\newblock In \emph{ECCV}, pages 275--291.

\bibitem[{Cheng et~al.(2016)Cheng, Xu, He, He, Wu, Sun, and
  Liu}]{cheng-etal-2016-semi}
Yong Cheng, Wei Xu, Zhongjun He, Wei He, Hua Wu, Maosong Sun, and Yang Liu.
  2016.
\newblock Semi-supervised learning for neural machine translation.
\newblock In \emph{Proceedings of the 54th Annual Meeting of the Association
  for Computational Linguistics (Volume 1: Long Papers)}, pages 1965--1974,
  Berlin, Germany. Association for Computational Linguistics.

\bibitem[{Clark et~al.(2018)Clark, Luong, Manning, and
  Le}]{clark2018semisupervised}
Kevin Clark, Minh-Thang Luong, Christopher~D. Manning, and Quoc~V. Le. 2018.
\newblock Semi-supervised sequence modeling with cross-view training.

\bibitem[{Cubuk et~al.(2018)Cubuk, Zoph, Mane, Vasudevan, and
  Le}]{cubuk2018autoaugment}
Ekin~D Cubuk, Barret Zoph, Dandelion Mane, Vijay Vasudevan, and Quoc~V Le.
  2018.
\newblock Autoaugment: Learning augmentation policies from data.
\newblock \emph{arXiv preprint arXiv:1805.09501}.

\bibitem[{Cubuk et~al.(2020)Cubuk, Zoph, Shlens, and Le}]{cubuk2020randaugment}
Ekin~D Cubuk, Barret Zoph, Jonathon Shlens, and Quoc~V Le. 2020.
\newblock Randaugment: Practical automated data augmentation with a reduced
  search space.
\newblock In \emph{Proceedings of the IEEE/CVF Conference on Computer Vision
  and Pattern Recognition Workshops}, pages 702--703.

\bibitem[{Dai and Le(2015)}]{dai2015semi}
Andrew~M Dai and Quoc~V Le. 2015.
\newblock Semi-supervised sequence learning.
\newblock In \emph{Advances in neural information processing systems}, pages
  3079--3087.

\bibitem[{Dai et~al.(2017)Dai, Yang, Yang, Cohen, and
  Salakhutdinov}]{dai2017good}
Zihang Dai, Zhilin Yang, Fan Yang, William~W. Cohen, and Ruslan Salakhutdinov.
  2017.
\newblock Good semi-supervised learning that requires a bad gan.

\bibitem[{Devlin et~al.(2018)Devlin, Chang, Lee, and
  Toutanova}]{devlin2018bert}
Jacob Devlin, Ming-Wei Chang, Kenton Lee, and Kristina Toutanova. 2018.
\newblock Bert: Pre-training of deep bidirectional transformers for language
  understanding.
\newblock \emph{arXiv preprint arXiv:1810.04805}.

\bibitem[{Edunov et~al.(2018)Edunov, Ott, Auli, and
  Grangier}]{edunov2018understanding}
Sergey Edunov, Myle Ott, Michael Auli, and David Grangier. 2018.
\newblock Understanding back-translation at scale.
\newblock \emph{arXiv preprint arXiv:1808.09381}.

\bibitem[{Fadaee et~al.(2017)Fadaee, Bisazza, and Monz}]{fadaee2017data}
Marzieh Fadaee, Arianna Bisazza, and Christof Monz. 2017.
\newblock Data augmentation for low-resource neural machine translation.
\newblock \emph{arXiv preprint arXiv:1705.00440}.

\bibitem[{Gao et~al.(2019)Gao, Zhu, Wu, Xia, Qin, Cheng, Zhou, and
  Liu}]{gao-etal-2019-soft}
Fei Gao, Jinhua Zhu, Lijun Wu, Yingce Xia, Tao Qin, Xueqi Cheng, Wengang Zhou,
  and Tie-Yan Liu. 2019.
\newblock Soft contextual data augmentation for neural machine translation.
\newblock In \emph{Proceedings of the 57th Annual Meeting of the Association
  for Computational Linguistics}, pages 5539--5544, Florence, Italy.
  Association for Computational Linguistics.

\bibitem[{Grandvalet and Bengio(2005)}]{grandvalet2005semi}
Yves Grandvalet and Yoshua Bengio. 2005.
\newblock Semi-supervised learning by entropy minimization.
\newblock In \emph{Advances in neural information processing systems}, pages
  529--536.

\bibitem[{Grill et~al.(2020)Grill, Strub, Altch{\'e}, Tallec, Richemond,
  Buchatskaya, Doersch, Pires, Guo, Azar et~al.}]{grill2020bootstrap}
Jean-Bastien Grill, Florian Strub, Florent Altch{\'e}, Corentin Tallec,
  Pierre~H Richemond, Elena Buchatskaya, Carl Doersch, Bernardo~Avila Pires,
  Zhaohan~Daniel Guo, Mohammad~Gheshlaghi Azar, et~al. 2020.
\newblock Bootstrap your own latent: A new approach to self-supervised
  learning.
\newblock \emph{arXiv preprint arXiv:2006.07733}.

\bibitem[{Guo et~al.(2019)Guo, Mao, and Zhang}]{guo2019augmenting}
Hongyu Guo, Yongyi Mao, and Richong Zhang. 2019.
\newblock Augmenting data with mixup for sentence classification: An empirical
  study.
\newblock \emph{arXiv preprint arXiv:1905.08941}.

\bibitem[{Gururangan et~al.(2020)Gururangan, Marasović, Swayamdipta, Lo,
  Beltagy, Downey, and Smith}]{dontstoppretraining2020}
Suchin Gururangan, Ana Marasović, Swabha Swayamdipta, Kyle Lo, Iz~Beltagy,
  Doug Downey, and Noah~A. Smith. 2020.
\newblock Don't stop pretraining: Adapt language models to domains and tasks.
\newblock In \emph{Proceedings of ACL}.

\bibitem[{Han et~al.(2018)Han, Yao, Yu, Niu, Xu, Hu, Tsang, and
  Sugiyama}]{han2018co}
Bo~Han, Quanming Yao, Xingrui Yu, Gang Niu, Miao Xu, Weihua Hu, Ivor Tsang, and
  Masashi Sugiyama. 2018.
\newblock Co-teaching: Robust training of deep neural networks with extremely
  noisy labels.
\newblock In \emph{Advances in neural information processing systems}, pages
  8527--8537.

\bibitem[{He et~al.(2019)He, Gu, Shen, and Ranzato}]{he2019revisiting}
Junxian He, Jiatao Gu, Jiajun Shen, and Marc'Aurelio Ranzato. 2019.
\newblock Revisiting self-training for neural sequence generation.
\newblock \emph{arXiv preprint arXiv:1909.13788}.

\bibitem[{Hochreiter and Schmidhuber(1997)}]{hochreiter1997long}
Sepp Hochreiter and J{\"u}rgen Schmidhuber. 1997.
\newblock Long short-term memory.
\newblock \emph{Neural computation}, 9(8):1735--1780.

\bibitem[{Howard and Ruder(2018)}]{howard2018universal}
Jeremy Howard and Sebastian Ruder. 2018.
\newblock Universal language model fine-tuning for text classification.
\newblock \emph{arXiv preprint arXiv:1801.06146}.

\bibitem[{Inoue(2018)}]{inoue2018data}
Hiroshi Inoue. 2018.
\newblock Data augmentation by pairing samples for images classification.
\newblock \emph{arXiv preprint arXiv:1801.02929}.

\bibitem[{Iscen et~al.(2019)Iscen, Tolias, Avrithis, and Chum}]{iscen2019label}
Ahmet Iscen, Giorgos Tolias, Yannis Avrithis, and Ondrej Chum. 2019.
\newblock Label propagation for deep semi-supervised learning.
\newblock In \emph{Proceedings of the IEEE conference on computer vision and
  pattern recognition}, pages 5070--5079.

\bibitem[{Joshi et~al.(2020)Joshi, Chen, Liu, Weld, Zettlemoyer, and
  Levy}]{danqi2020spanbert}
Mandar Joshi, Danqi Chen, Yinhan Liu, Daniel~S Weld, Luke Zettlemoyer, and Omer
  Levy. 2020.
\newblock Spanbert: Improving pre-training by representing and predicting
  spans.
\newblock \emph{Transactions of the Association for Computational Linguistics},
  8:64--77.

\bibitem[{Karamanolakis et~al.(2019)Karamanolakis, Hsu, and
  Gravano}]{karamanolakis2019leveraging}
Giannis Karamanolakis, Daniel Hsu, and Luis Gravano. 2019.
\newblock Leveraging just a few keywords for fine-grained aspect detection
  through weakly supervised co-training.

\bibitem[{Kim(2014)}]{kim-2014-convolutional}
Yoon Kim. 2014.
\newblock Convolutional neural networks for sentence classification.
\newblock In \emph{Proceedings of the 2014 Conference on Empirical Methods in
  Natural Language Processing ({EMNLP})}, pages 1746--1751, Doha, Qatar.
  Association for Computational Linguistics.

\bibitem[{Kingma and Ba(2014)}]{kingma2014adam}
Diederik~P Kingma and Jimmy Ba. 2014.
\newblock Adam: A method for stochastic optimization.
\newblock \emph{arXiv preprint arXiv:1412.6980}.

\bibitem[{Kobayashi(2018)}]{kobayashi2018contextual}
Sosuke Kobayashi. 2018.
\newblock Contextual augmentation: Data augmentation by words with paradigmatic
  relations.
\newblock \emph{arXiv preprint arXiv:1805.06201}.

\bibitem[{Krizhevsky et~al.(2012)Krizhevsky, Sutskever, and
  Hinton}]{krizhevsky2012imagenet}
Alex Krizhevsky, Ilya Sutskever, and Geoffrey~E Hinton. 2012.
\newblock Imagenet classification with deep convolutional neural networks.
\newblock In \emph{Advances in neural information processing systems}, pages
  1097--1105.

\bibitem[{Kumar et~al.(2020)Kumar, Choudhary, and Cho}]{kumar2020data}
Varun Kumar, Ashutosh Choudhary, and Eunah Cho. 2020.
\newblock Data augmentation using pre-trained transformer models.

\bibitem[{Laine and Aila(2016)}]{laine2016temporal}
Samuli Laine and Timo Aila. 2016.
\newblock Temporal ensembling for semi-supervised learning.
\newblock \emph{arXiv preprint arXiv:1610.02242}.

\bibitem[{Laine and Aila(2017)}]{laine2017temporal}
Samuli Laine and Timo Aila. 2017.
\newblock Temporal ensembling for semi-supervised learning.

\bibitem[{Lee(2013)}]{lee2013pseudo}
Dong-Hyun Lee. 2013.
\newblock Pseudo-label: The simple and efficient semi-supervised learning
  method for deep neural networks.
\newblock In \emph{Workshop on challenges in representation learning, ICML},
  volume~3.

\bibitem[{Levy and Goldberg(2014)}]{wordvec2014matrixfactor}
Omer Levy and Yoav Goldberg. 2014.
\newblock Neural word embedding as implicit matrix factorization.
\newblock In \emph{Advances in Neural Information Processing Systems 27: Annual
  Conference on Neural Information Processing Systems 2014, December 8-13 2014,
  Montreal, Quebec, Canada}, pages 2177--2185.

\bibitem[{Lewis et~al.(2019)Lewis, Liu, Goyal, Ghazvininejad, Mohamed, Levy,
  Stoyanov, and Zettlemoyer}]{lewis2019bart}
Mike Lewis, Yinhan Liu, Naman Goyal, Marjan Ghazvininejad, Abdelrahman Mohamed,
  Omer Levy, Ves Stoyanov, and Luke Zettlemoyer. 2019.
\newblock Bart: Denoising sequence-to-sequence pre-training for natural
  language generation, translation, and comprehension.
\newblock \emph{arXiv preprint arXiv:1910.13461}.

\bibitem[{Li et~al.(2014)Li, Ott, Cardie, and Hovy}]{li-etal-2014-towards}
Jiwei Li, Myle Ott, Claire Cardie, and Eduard Hovy. 2014.
\newblock Towards a general rule for identifying deceptive opinion spam.
\newblock In \emph{Proceedings of the 52nd Annual Meeting of the Association
  for Computational Linguistics (Volume 1: Long Papers)}, pages 1566--1576,
  Baltimore, Maryland. Association for Computational Linguistics.

\bibitem[{Li et~al.(2019{\natexlab{a}})Li, Sun, Liu, Zheng, Zhou, Chua, and
  Schiele}]{li2019learning}
Xinzhe Li, Qianru Sun, Yaoyao Liu, Shibao Zheng, Qin Zhou, Tat-Seng Chua, and
  Bernt Schiele. 2019{\natexlab{a}}.
\newblock Learning to self-train for semi-supervised few-shot classification.

\bibitem[{Li et~al.(2019{\natexlab{b}})Li, Liu, and Tan}]{li2019decoupled}
Yiting Li, Lu~Liu, and Robby~T Tan. 2019{\natexlab{b}}.
\newblock Decoupled certainty-driven consistency loss for semi-supervised
  learning.
\newblock \emph{arXiv preprint arXiv:1901.05657}.

\bibitem[{Liao and Veeramachaneni(2009)}]{liao2009simple}
Wenhui Liao and Sriharsha Veeramachaneni. 2009.
\newblock A simple semi-supervised algorithm for named entity recognition.
\newblock In \emph{Proceedings of the NAACL HLT 2009 Workshop on
  Semi-Supervised Learning for Natural Language Processing}, pages 58--65.

\bibitem[{Liu et~al.(2020)Liu, Li, and Tan}]{liu2020decoupled}
Lu~Liu, Yiting Li, and Robby~T. Tan. 2020.
\newblock Decoupled certainty-driven consistency loss for semi-supervised
  learning.

\bibitem[{Liu et~al.(2019)Liu, Ott, Goyal, Du, Joshi, Chen, Levy, Lewis,
  Zettlemoyer, and Stoyanov}]{yinhan2019roberta}
Yinhan Liu, Myle Ott, Naman Goyal, Jingfei Du, Mandar Joshi, Danqi Chen, Omer
  Levy, Mike Lewis, Luke Zettlemoyer, and Veselin Stoyanov. 2019.
\newblock Roberta: A robustly optimized bert pretraining approach.
\newblock \emph{arXiv preprint arXiv:1907.11692}.

\bibitem[{Luo et~al.(2018)Luo, Zhu, Li, Ren, and Zhang}]{luo2018smooth}
Yucen Luo, Jun Zhu, Mengxi Li, Yong Ren, and Bo~Zhang. 2018.
\newblock Smooth neighbors on teacher graphs for semi-supervised learning.

\bibitem[{Maas et~al.(2011)Maas, Daly, Pham, Huang, Ng, and
  Potts}]{maas2011learning}
Andrew~L Maas, Raymond~E Daly, Peter~T Pham, Dan Huang, Andrew~Y Ng, and
  Christopher Potts. 2011.
\newblock Learning word vectors for sentiment analysis.
\newblock In \emph{Proceedings of the 49th annual meeting of the association
  for computational linguistics: Human language technologies-volume 1}, pages
  142--150. Association for Computational Linguistics.

\bibitem[{Mikolov et~al.(2013{\natexlab{a}})Mikolov, Chen, Corrado, and
  Dean}]{mikolov2013efficient}
Tomas Mikolov, Kai Chen, Greg Corrado, and Jeffrey Dean. 2013{\natexlab{a}}.
\newblock Efficient estimation of word representations in vector space.
\newblock \emph{arXiv preprint arXiv:1301.3781}.

\bibitem[{Mikolov et~al.(2013{\natexlab{b}})Mikolov, Sutskever, Chen, Corrado,
  and Dean}]{mikolov2013distributed}
Tomas Mikolov, Ilya Sutskever, Kai Chen, Greg~S Corrado, and Jeff Dean.
  2013{\natexlab{b}}.
\newblock Distributed representations of words and phrases and their
  compositionality.
\newblock In \emph{Advances in neural information processing systems}, pages
  3111--3119.

\bibitem[{Miyato et~al.(2016)Miyato, Dai, and
  Goodfellow}]{miyato2016adversarial}
Takeru Miyato, Andrew~M Dai, and Ian Goodfellow. 2016.
\newblock Adversarial training methods for semi-supervised text classification.
\newblock \emph{arXiv preprint arXiv:1605.07725}.

\bibitem[{Miyato et~al.(2018)Miyato, ichi Maeda, Koyama, and
  Ishii}]{miyato2018virtual}
Takeru Miyato, Shin ichi Maeda, Masanori Koyama, and Shin Ishii. 2018.
\newblock Virtual adversarial training: A regularization method for supervised
  and semi-supervised learning.

\bibitem[{Nigam et~al.(2006)Nigam, McCallum, and Mitchell}]{nigam2006semi}
Kamal Nigam, Andrew McCallum, and Tom~M Mitchell. 2006.
\newblock Semi-supervised text classification using em.

\bibitem[{Ott et~al.(2011)Ott, Choi, Cardie, and
  Hancock}]{ott-etal-2011-finding}
Myle Ott, Yejin Choi, Claire Cardie, and Jeffrey~T. Hancock. 2011.
\newblock Finding deceptive opinion spam by any stretch of the imagination.
\newblock In \emph{Proceedings of the 49th Annual Meeting of the Association
  for Computational Linguistics: Human Language Technologies}, pages 309--319,
  Portland, Oregon, USA. Association for Computational Linguistics.

\bibitem[{Park et~al.(2020)Park, Zhang, Jia, Han, Chiu, Li, Wu, and
  Le}]{park2020improved}
Daniel~S Park, Yu~Zhang, Ye~Jia, Wei Han, Chung-Cheng Chiu, Bo~Li, Yonghui Wu,
  and Quoc~V Le. 2020.
\newblock Improved noisy student training for automatic speech recognition.
\newblock \emph{arXiv preprint arXiv:2005.09629}.

\bibitem[{Parthasarathi and Strom(2019)}]{parthasarathi2019lessons}
Sree Hari~Krishnan Parthasarathi and Nikko Strom. 2019.
\newblock Lessons from building acoustic models with a million hours of speech.

\bibitem[{Paulin et~al.(2014)Paulin, Revaud, Harchaoui, Perronnin, and
  Schmid}]{paulin2014transformation}
Mattis Paulin, J{\'e}r{\^o}me Revaud, Zaid Harchaoui, Florent Perronnin, and
  Cordelia Schmid. 2014.
\newblock Transformation pursuit for image classification.
\newblock In \emph{Proceedings of the IEEE conference on computer vision and
  pattern recognition}, pages 3646--3653.

\bibitem[{Pennington et~al.(2014)Pennington, Socher, and
  Manning}]{pennington2014glove}
Jeffrey Pennington, Richard Socher, and Christopher~D Manning. 2014.
\newblock Glove: Global vectors for word representation.
\newblock In \emph{Proceedings of the 2014 conference on empirical methods in
  natural language processing (EMNLP)}, pages 1532--1543.

\bibitem[{Peters et~al.(2017)Peters, Ammar, Bhagavatula, and
  Power}]{peters2017semisupervised}
Matthew~E. Peters, Waleed Ammar, Chandra Bhagavatula, and Russell Power. 2017.
\newblock Semi-supervised sequence tagging with bidirectional language models.

\bibitem[{Qader et~al.(2019)Qader, Portet, and
  Labbé}]{qader2019semisupervised}
Raheel Qader, François Portet, and Cyril Labbé. 2019.
\newblock Semi-supervised neural text generation by joint learning of natural
  language generation and natural language understanding models.

\bibitem[{Qiao et~al.(2018)Qiao, Shen, Zhang, Wang, and Yuille}]{qiao2018deep}
Siyuan Qiao, Wei Shen, Zhishuai Zhang, Bo~Wang, and Alan Yuille. 2018.
\newblock Deep co-training for semi-supervised image recognition.

\bibitem[{Ramachandran et~al.(2016)Ramachandran, Liu, and
  Le}]{ramachandran2016unsupervised}
Prajit Ramachandran, Peter~J Liu, and Quoc~V Le. 2016.
\newblock Unsupervised pretraining for sequence to sequence learning.
\newblock \emph{arXiv preprint arXiv:1611.02683}.

\bibitem[{Rasmus et~al.(2015)Rasmus, Valpola, Honkala, Berglund, and
  Raiko}]{rasmus2015semisupervised}
Antti Rasmus, Harri Valpola, Mikko Honkala, Mathias Berglund, and Tapani Raiko.
  2015.
\newblock Semi-supervised learning with ladder networks.

\bibitem[{Reed et~al.(2015)Reed, Lee, Anguelov, Szegedy, Erhan, and
  Rabinovich}]{reed2015training}
Scott Reed, Honglak Lee, Dragomir Anguelov, Christian Szegedy, Dumitru Erhan,
  and Andrew Rabinovich. 2015.
\newblock Training deep neural networks on noisy labels with bootstrapping.

\bibitem[{Riloff and Wiebe(2003)}]{riloff2003learning}
Ellen Riloff and Janyce Wiebe. 2003.
\newblock Learning extraction patterns for subjective expressions.
\newblock In \emph{Proceedings of the 2003 conference on Empirical methods in
  natural language processing}, pages 105--112.

\bibitem[{{\c{S}}ahin and Steedman(2018)}]{sahin-steedman-2018-data}
G{\"o}zde~G{\"u}l {\c{S}}ahin and Mark Steedman. 2018.
\newblock Data augmentation via dependency tree morphing for low-resource
  languages.
\newblock In \emph{Proceedings of the 2018 Conference on Empirical Methods in
  Natural Language Processing}, pages 5004--5009, Brussels, Belgium.
  Association for Computational Linguistics.

\bibitem[{Sajjadi et~al.(2016)Sajjadi, Javanmardi, and
  Tasdizen}]{sajjadi2016regularization}
Mehdi Sajjadi, Mehran Javanmardi, and Tolga Tasdizen. 2016.
\newblock Regularization with stochastic transformations and perturbations for
  deep semi-supervised learning.
\newblock In \emph{Advances in neural information processing systems}, pages
  1163--1171.

\bibitem[{{Scudder}(1965)}]{1053799}
H.~{Scudder}. 1965.
\newblock Probability of error of some adaptive pattern-recognition machines.
\newblock \emph{IEEE Transactions on Information Theory}, 11(3):363--371.

\bibitem[{Sennrich et~al.(2016)Sennrich, Haddow, and
  Birch}]{sennrich2016back-translation}
Rico Sennrich, Barry Haddow, and Alexandra Birch. 2016.
\newblock Improving neural machine translation models with monolingual data.
\newblock In \emph{Proceedings of the 54th Annual Meeting of the Association
  for Computational Linguistics (Volume 1: Long Papers)}, pages 86--96, Berlin,
  Germany.

\bibitem[{Shang et~al.(2019)Shang, Li, Fu, Bing, Zhao, Shi, and
  Yan}]{shang2019semisupervised}
Mingyue Shang, Piji Li, Zhenxin Fu, Lidong Bing, Dongyan Zhao, Shuming Shi, and
  Rui Yan. 2019.
\newblock Semi-supervised text style transfer: Cross projection in latent
  space.

\bibitem[{Shi et~al.(2018)Shi, Gong, Ding, Tao, and Zheng}]{Shi_2018_ECCV}
Weiwei Shi, Yihong Gong, Chris Ding, Zhiheng~MaXiaoyu Tao, and Nanning Zheng.
  2018.
\newblock Transductive semi-supervised deep learning using min-max features.
\newblock In \emph{Proceedings of the European Conference on Computer Vision
  (ECCV)}.

\bibitem[{Shleifer(2019)}]{shleifer2019low}
Sam Shleifer. 2019.
\newblock Low resource text classification with ulmfit and backtranslation.

\bibitem[{Singh et~al.(2019)Singh, McCann, Keskar, Xiong, and
  Socher}]{singh2019xlda}
Jasdeep Singh, Bryan McCann, Nitish~Shirish Keskar, Caiming Xiong, and Richard
  Socher. 2019.
\newblock Xlda: Cross-lingual data augmentation for natural language inference
  and question answering.

\bibitem[{Tan and Le(2020)}]{tan2020efficientnet}
Mingxing Tan and Quoc~V. Le. 2020.
\newblock Efficientnet: Rethinking model scaling for convolutional neural
  networks.

\bibitem[{Tang et~al.(2016)Tang, Qin, Feng, and Liu}]{tang-etal-2016-effective}
Duyu Tang, Bing Qin, Xiaocheng Feng, and Ting Liu. 2016.
\newblock Effective {LSTM}s for target-dependent sentiment classification.
\newblock In \emph{Proceedings of {COLING} 2016, the 26th International
  Conference on Computational Linguistics: Technical Papers}, pages 3298--3307,
  Osaka, Japan. The COLING 2016 Organizing Committee.

\bibitem[{Tarvainen and Valpola(2018)}]{tarvainen2018mean}
Antti Tarvainen and Harri Valpola. 2018.
\newblock Mean teachers are better role models: Weight-averaged consistency
  targets improve semi-supervised deep learning results.

\bibitem[{Tu et~al.(2016)Tu, Liu, Shang, Liu, and Li}]{tu2016neural}
Zhaopeng Tu, Yang Liu, Lifeng Shang, Xiaohua Liu, and Hang Li. 2016.
\newblock Neural machine translation with reconstruction.
\newblock \emph{arXiv preprint arXiv:1611.01874}.

\bibitem[{Verma et~al.(2019)Verma, Lamb, Kannala, Bengio, and
  Lopez-Paz}]{verma2019interpolation}
Vikas Verma, Alex Lamb, Juho Kannala, Yoshua Bengio, and David Lopez-Paz. 2019.
\newblock Interpolation consistency training for semi-supervised learning.

\bibitem[{Wang and Yang(2015)}]{wang-yang-2015-thats}
William~Yang Wang and Diyi Yang. 2015.
\newblock That{'}s so annoying!!!: A lexical and frame-semantic embedding based
  data augmentation approach to automatic categorization of annoying behaviors
  using {\#}petpeeve tweets.
\newblock In \emph{Proceedings of the 2015 Conference on Empirical Methods in
  Natural Language Processing}, pages 2557--2563, Lisbon, Portugal. Association
  for Computational Linguistics.

\bibitem[{Wei and Zou(2019)}]{wei2019eda}
Jason Wei and Kai Zou. 2019.
\newblock Eda: Easy data augmentation techniques for boosting performance on
  text classification tasks.

\bibitem[{Wu et~al.(2019)Wu, Lv, Zang, Han, and Hu}]{wu2019conditional}
Xing Wu, Shangwen Lv, Liangjun Zang, Jizhong Han, and Songlin Hu. 2019.
\newblock Conditional bert contextual augmentation.
\newblock In \emph{International Conference on Computational Science}, pages
  84--95. Springer.

\bibitem[{Xia et~al.(2019)Xia, Kong, Anastasopoulos, and
  Neubig}]{xia-etal-2019-generalized}
Mengzhou Xia, Xiang Kong, Antonios Anastasopoulos, and Graham Neubig. 2019.
\newblock Generalized data augmentation for low-resource translation.
\newblock In \emph{Proceedings of the 57th Annual Meeting of the Association
  for Computational Linguistics}, pages 5786--5796, Florence, Italy.
  Association for Computational Linguistics.

\bibitem[{Xie et~al.(2019)Xie, Dai, Hovy, Luong, and Le}]{xie2019unsupervised}
Qizhe Xie, Zihang Dai, Eduard Hovy, Minh-Thang Luong, and Quoc~V Le. 2019.
\newblock Unsupervised data augmentation for consistency training.
\newblock \emph{arXiv preprint arXiv:1904.12848}.

\bibitem[{Xie et~al.(2020)Xie, Luong, Hovy, and Le}]{xie2020self}
Qizhe Xie, Minh-Thang Luong, Eduard Hovy, and Quoc~V Le. 2020.
\newblock Self-training with noisy student improves imagenet classification.
\newblock In \emph{Proceedings of the IEEE/CVF Conference on Computer Vision
  and Pattern Recognition}, pages 10687--10698.

\bibitem[{Xie et~al.(2018)Xie, Genthial, Xie, Ng, and
  Jurafsky}]{xie2018noising}
Ziang Xie, Guillaume Genthial, Stanley Xie, Andrew~Y Ng, and Dan Jurafsky.
  2018.
\newblock Noising and denoising natural language: Diverse backtranslation for
  grammar correction.
\newblock In \emph{Proceedings of the 2018 Conference of the North American
  Chapter of the Association for Computational Linguistics: Human Language
  Technologies, Volume 1 (Long Papers)}, pages 619--628.

\bibitem[{Xie et~al.(2017)Xie, Wang, Li, L{\'e}vy, Nie, Jurafsky, and
  Ng}]{xie2017data}
Ziang Xie, Sida~I Wang, Jiwei Li, Daniel L{\'e}vy, Aiming Nie, Dan Jurafsky,
  and Andrew~Y Ng. 2017.
\newblock Data noising as smoothing in neural network language models.
\newblock \emph{arXiv preprint arXiv:1703.02573}.

\bibitem[{Yalniz et~al.(2019)Yalniz, J{\'e}gou, Chen, Paluri, and
  Mahajan}]{yalniz2019billion}
I~Zeki Yalniz, Herv{\'e} J{\'e}gou, Kan Chen, Manohar Paluri, and Dhruv
  Mahajan. 2019.
\newblock Billion-scale semi-supervised learning for image classification.
\newblock \emph{arXiv preprint arXiv:1905.00546}.

\bibitem[{Yang et~al.(2019)Yang, Dai, Yang, Carbonell, Salakhutdinov, and
  Le}]{yang2019xlnet}
Zhilin Yang, Zihang Dai, Yiming Yang, Jaime Carbonell, Russ~R Salakhutdinov,
  and Quoc~V Le. 2019.
\newblock Xlnet: Generalized autoregressive pretraining for language
  understanding.
\newblock In \emph{Advances in neural information processing systems}, pages
  5753--5763.

\bibitem[{Zang and Wan(2019)}]{zang2019semisupervised}
Hongyu Zang and Xiaojun Wan. 2019.
\newblock A semi-supervised approach for low-resourced text generation.

\bibitem[{Zhang et~al.(2015)Zhang, Zhao, and LeCun}]{zhang2015character}
Xiang Zhang, Junbo Zhao, and Yann LeCun. 2015.
\newblock Character-level convolutional networks for text classification.
\newblock In \emph{Advances in neural information processing systems}, pages
  649--657.

\bibitem[{Zhu(2005)}]{zhu2005semi}
Xiaojin~Jerry Zhu. 2005.
\newblock Semi-supervised learning literature survey.
\newblock Technical report, University of Wisconsin-Madison Department of
  Computer Sciences.

\bibitem[{Zoph et~al.(2020)Zoph, Ghiasi, Lin, Cui, Liu, Cubuk, and
  Le}]{zoph2020rethinking}
Barret Zoph, Golnaz Ghiasi, Tsung-Yi Lin, Yin Cui, Hanxiao Liu, Ekin~D Cubuk,
  and Quoc~V Le. 2020.
\newblock Rethinking pre-training and self-training.
\newblock \emph{arXiv preprint arXiv:2006.06882}.

\end{thebibliography}
\bibliographystyle{acl_natbib}

\end{document}